\documentclass[letterpaper, 10 pt, conference]{ieeeconf}  

\IEEEoverridecommandlockouts                              

\usepackage{graphics} 
\usepackage{epsfig} 
\usepackage{mathptmx} 
\usepackage{times} 
\usepackage{amsmath} 
\newcommand{\norm}[1]{\left\lVert#1\right\rVert}
\usepackage{amssymb}  
\usepackage{soul}

\usepackage{fancyhdr,graphicx,amsmath,amssymb,mathtools}
\usepackage[ruled,vlined]{algorithm2e}
\usepackage{gensymb}
\usepackage{booktabs}
\usepackage{pifont}
\usepackage{url}
\usepackage{breakurl}
\usepackage[breaklinks]{hyperref}
\usepackage{forest}

\usepackage{caption}
\usepackage{subcaption}
\PassOptionsToPackage{hyphens}{url}\usepackage{hyperref}

\title{\LARGE \bf
Optical Flow based Visual Potential Field for Autonomous Driving}

\author{
Linda Capito$^*$, Umit Ozguner$^*$, Keith Redmill$^*$
\thanks{*The authors are with the Department of Electrical and Computer Engineering at Ohio State University, OH, USA} \\
\thanks{Corresponding author: Linda Capito, capitoruiz.1@osu.edu}}

\begin{document}

\maketitle
\thispagestyle{empty}
\pagestyle{empty}

\begin{abstract}

Monocular vision based navigation for automated driving is a challenging task due to the lack of enough information to compute temporal relationships among objects on the road. Optical flow is an option to obtain temporal information from monocular camera images, and has been used widely with the purpose of identifying objects and their relative motion. This work proposes to generate an artificial potential field, i.e. visual potential field, from a sequence of images using sparse optical flow, which is used together with a gradient tracking sliding mode controller to navigate the vehicle to destination without collision with obstacles. The angular reference for the vehicle is computed online. This work considers that the vehicle does not require to have \textit{a priori} information from the map or obstacles to navigate successfully. The proposed technique is tested both in synthetic and real images.




\end{abstract}

\section{INTRODUCTION}

Vision based vehicular navigation with a single camera has been shown to be a challenging problem within the field of computer vision, mainly because it is hard to obtain a force to control the vehicle just from monocular images \cite{bonin2008visual,ohnishi2008visual}. To obtain this force, we can compute a so called visual potential field, which is an approximation of the projection of the potential field in the 3D world to the image plane. This technique has been used with success for Unmanned Aerial Vehicles (UAVs) \cite{miao2017optical} and it has also been applied to robots, in environments which are basically restriction free, i.e. the robot can move to whichever direction it finds suitable \cite{huang2006visual}. 

Navigation for cars is different from UAVs and mobile robots in the sense that a car has additional constraints from the road, like lane boundaries. Then, it is necessary to create those restrictions from the available visual information. One way to get the visual potential field is to adequately estimate the optical flow from the scene.

Optical flow is a vector field that consists of the direction and magnitude of color intensity changes from the movement of objects with the same brightness value or feature pattern between two consequent images, obtained from the projection of an object in a 3D space onto an image plane. So, when an object moves, its projection will change position in the image plane and generate several vectors (direction, magnitude) \cite{camus1995calculating}. This allows to infer temporal information about the current scene, while being less computationally expensive than processing full raw images, which makes it an attractive method for online navigation controllers. 

Optical flow has been widely used for obstacle avoidance and vehicular/mobile robot navigation. For instance, optical flow was used to recognize the lane boundaries of the road under adverse weather conditions as reported by \cite{gern2002vision}. In \cite{lieb2005adaptive} this technique was used within a road following algorithm that allowed to identify the road without making assumptions about its structure or appearance. It was used to get a steering angle through getting the optical flow between subsequent images in \cite{yoshimoto1995automatic}. More recently, optical flow has been used as a mean to model human behavior, as in \cite{okafuji2015development}, where an optical flow automatic steering system for the vehicle is presented.

Referring specifically to direct control of the vehicle, the most common idea is to estimate the time to collision (TTC) from the optical flow vector field and use it to steer away from static or dynamic obstacles as seen in \cite{camus1995calculating}. There is also the balance strategy approach, which consists of a set of behavior rules to be applied when the average of the vector field is above/below certain predefined thresholds \cite{souhila2007optical,chang2016experimental}. In \cite{ohnishi2008visual}, the authors get an visual potential field from the optical flow vector field by identifying the dominant plane on the image. Then apply a balance strategy for the robot navigation. 

We propose to compute an Artificial Potential Field APF (Visual Potential Field) from the optical flow vector field information, that contains both the information from the obstacles as the restrictions of the road. In general, an APF  generates a surface where the target is a global minimum and the obstacles are local maxima. The target generates an attractive force on the vehicle while obstacles create a repulsive force. The gradient lines from the optical flow are used to generate this field and get the navigation reference \cite{guldner1995tracking,ferrara2007gradient}.

To include the constraints from the road, a \textit{road potential field} can be used, such as in \cite{snapper2018model}, where the field provides a steep slope when it is closer to the boundaries of the road and has local minima along the center of the lane. Other approaches are introduced in \cite{hamid2018,raksincharoensak2016vehicle}, where the distance to the lane boundaries and obstacles are used to compute a total risk field $U_{risk}$. These contributions consider certain knowledge about the geometry of the road.


Unlike \cite{ohnishi2008visual,miao2017optical}, the presented approach works for automated vehicles that have to navigate in a restricted environment (i.e. follow the road), hence the contribution of this work lies in combining previous techniques proven in mobile/aerial robots and extending them to work with cars, and is based on \cite{capito2019optical}.

The paper is organized as follows: Section II presents the problem formulation, then Section III introduces Optical Flow and focus of expansion (FOE) definitions. Section IV presents the computation of the vision potential field based on optical flow and Section V introduces the controller design. Section VI presents evaluation of the proposed method under different weather conditions in two sets of images. Finally, conclusions are presented in Section VII.

\section{Problem formulation}

\begin{figure}[h!]
    \centering
    \includegraphics[width=\linewidth]{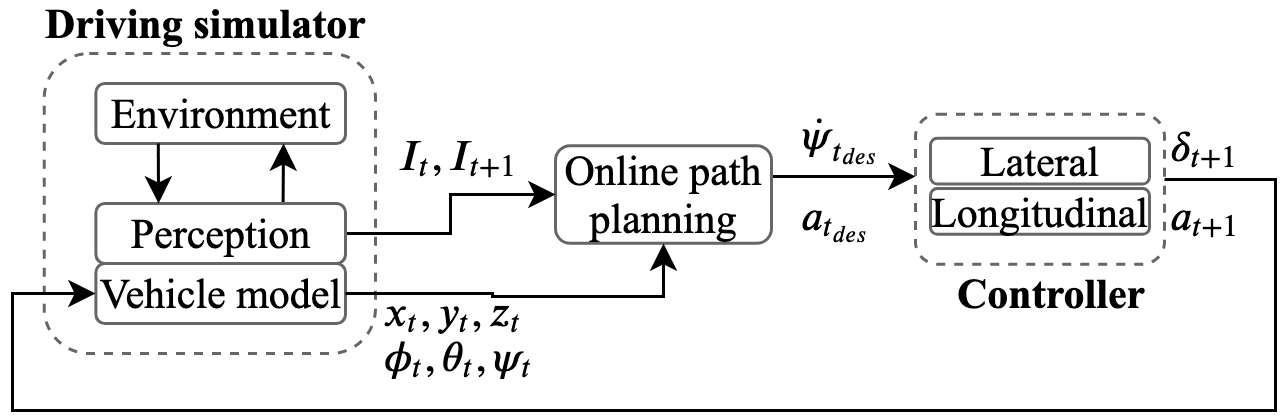}
    \caption{Proposed control framework}
    \label{fig:basicdiagram}
    \end{figure}

The problem that we solve is the visual based navigation of an autonomous vehicle using a monocular camera. Fig.\ref{fig:basicdiagram} presents the modules considered for the implementation of the solution. 

It is considered that we have access to all the ground truth data from the vehicle, position ($x,y,z$) and orientation (roll $\phi$, pitch $\theta$, yaw $\psi$). The vehicle has a monocular RGB camera mounted on the windshield, at a fixed position ($x_c,y_c,z_c$) with respect to the center of mass of the vehicle. The camera provides images ($I_t$) at each timestep. The \textit{Online Path Planning} block takes pairs of images ($I_t, I_{t+1}$) and computes the optical flow vector field, the Focus of Expansion (FOE) and the visual potential field, which allows to generate the speed and heading reference, which are sent to the \textit{Controller} block. This last block computes the acceleration ($a$) and steering ($\delta$) control actions using a gradient tracking sliding mode controller \cite{guldner1995tracking}. The control actions are then sent to the vehicle. 

\section{Optical Flow}

Given a set of pixels on an image $I_t$, the optical flow vector field can find those same pixels in the subsequent image $I_{t+1}$. 



Given two subsequent images, they are first transformed to grayscale, then the corners are detected using the Shi-Tomasi algorithm \cite{shi1993good}, and finally, the algorithm Lucas-Kanade (LK) with Pyramidal implementation \cite{bouguet2001pyramidal} is employed to obtain the optical flow vector field.


\subsection{Focus of Expansion (FOE)}

The Focus of Expansion (FOE) point shows the direction of the vehicle motion, and is computed as the intersection of all the optical flow vectors \cite{okafuji2015development}.

The FOE is obtained through a Least-Squares formulation \cite{okafuji2015development}, where we minimize the error with respect to the gradient of the optical flow. Consider image $I$ has pixels $p_i(x,y),i=1,...,n$, where $(x,y)$ is the position of the pixel $p_i$ in the image plane. We can build the matrices $A$ and $b$ using the spatial gradient $\nabla I(p_i)$ and the temporal gradient $I_t(p_i)$ respectively:

\begin{align} \label{eq:A}
\centering
    & A\quad =\quad \left[ \begin{matrix} \nabla I(p_{ 1 }) \\ \nabla I(p_{ 2 }) \\ \vdots  \\ \nabla I(p_{ n }) \end{matrix} \right] = \left[ \begin{matrix} a_{00} & a_{01}\\ a_{10} & a_{11} \\ \vdots & \vdots \\ a_{n0} & a_{n1} \end{matrix} \right] \quad \quad \quad \\
    & b\quad =\quad \left[ \begin{matrix} -I_{ t }(p_{ 1 }) \\ -I_{ t }(p_{ 2 }) \\ \vdots  \\ -I_{ t }(p_{ n }) \end{matrix} \right] = \left[ \begin{matrix}  b_0\\ b_1 \\ \vdots \\ b_n  \end{matrix} \right]
\end{align}

For each pixel $p_i(x,y)$ the obtained flow vector $(v_x,v_y)$ provides $a_{i0}=v_y$, $a_{i1}=-v_x$ and $b_i=xv_y-yv_x$. Then the location of the FOE $(x_{FOE},y_{FOE})$ on the image plane is given by:

\begin{flalign*}
    &FOE = (A^TA)^{-1}A^Tb\\
    &=\left[ \begin{matrix} \sum { a_{ i0 }b_{ i } } \sum { a_{ j1 }^{ 2 } } -\sum { a_{ i1 }b_{ i } } \sum { a_{ j0 }a_{ j1 } }  \\ -\sum { a_{ i0 }b_{ i } } \sum { a_{ j0 }a_{ j1 } } +\sum { a_{ i1 }b_{ i } } \sum { a_{ j0 }^{ 2 } }  \end{matrix} \right] -\frac { 1 }{ \sum { a_{ j0 }^{ 2 }a_{ j1 }^{ 2 } } -(\sum{a_{i0}a_{i1}})^{ 2 } }   
\end{flalign*}
    
\subsection{Depth from Optical Flow}
    
To recover depth from the most important pixels on the image is possible \cite{capito2019optical} but very inaccurate, as the optical flow values are noisy near the FOE. Hence, we use instead the time to contact (TTC), which is depth in terms of time and provides enough information about the vehicle motion. The TTC for each pixel $p_i(x,y)$ with optical flow vector field $(v_x,v_y)$ is computed as follows: 

\begin{equation} \label{eq:TTC}
    TTC_i = \frac{\sqrt{(x-x_{FOE})^2+(y-y_{FOE})^2}}{\sqrt{v_x^2+v_y^2}}
\end{equation}    

\subsection{Obstacle detection}\label{sec:obs_det}

The optical flow vector field presents a disturbance when an obstacle is present in the image. While the disturbance may be heuristically computed, a feasible method to distinguish the background from the obstacles using the optical flow field is to use the Otsu threshold segmentation method \cite{otsu1979threshold}.

Let $I(x,y,t):=I_t(x,y)$ be the image at time $t$. Let $O(x,y,t):=O_t(x,y)$ be the plane that shows only the obstacles in the image obtained by the LK algorithm, i.e. a binary image \cite{ohnishi2008visual}. Then to get the gradient, we use a smoothing Gaussian function $G(x,y)$:

\begin{equation} \label{eq:G}
    G(x,y) = \frac{1}{2 \pi \sigma} e^{-\frac{x^2+y^2}{2\sigma ^2}}
\end{equation}

The parameter $\sigma$ can be chosen as half the image width/length. Then, we get the convolution between $O(x,y,t)$ and \eqref{eq:G}:

\begin{equation}
    G*O(x,y,t)= \int_{- \infty}^{\infty} \int_{- \infty}^{\infty} G(u-x,v-y)O(x,y,t)du dv
\end{equation}

Which allows to obtain a function $g(x,y,t)$:

\begin{equation} \label{eq:g}
    g(x,y,t)= \nabla(G * O(x,y,t))= \bigg( \begin{matrix} \frac{\partial}{\partial x} (G * O(x,y,t)) \\
    \frac{\partial}{\partial y} (G * O(x,y,t))
    \end{matrix} \bigg)
\end{equation}

Where Eq. \eqref{eq:g} represents the gradient vector of $O_t(x,y)$.

\section{Visual Potential Field}

We compute separately the potential fields for the target, the obstacles, and for the road, and then sum them up to get the total potential field at a certain time $t$.

\subsection{Target Potential Field} \label{sec:tpf}

This potential field is directly proportional to the Euclidean distance away from the vehicle, so can be obtained by:

\begin{equation}
    U_{att}=\frac{1}{2}\alpha \norm{\sqrt{(x-x_{goal})^2+(y-y_{goal})^2}}
\end{equation}

The goal term pulls the vehicle towards the goal. Its strength increases proportionally with the distance to the goal and its adjusted through the constant $\alpha$. The module of the attraction force is the gradient of the potential field:

\begin{equation}
    F_{att}=\nabla U_{att}=\alpha\sqrt{(x-x_{goal})^2+(y-y_{goal})^2}
\end{equation}

The angle that this force makes with the image plane is equal to the goal angle, i.e. $\theta_{goal}=atan \frac{y-y_{goal}}{x-x_{goal}}$. 

\subsection{Obstacle Potential Field} \label{sec:opf}

The obstacle potential field is an approximation of the projection of the 3D potential field onto the image plane, hence the gradient vector $g(x,y,t)$ of $O_t(x,y)$ found in Section \ref{sec:obs_det}. The TTC for each pixel in the image is also used for the definition of the repulsive force \cite{ohnishi2008visual,miao2017optical}:

\begin{equation}
    F_{rep} = \bigg( \begin{matrix} F_x\\ F_y \end{matrix} \bigg) = \gamma \frac{1}{|R|} \Bigg( \begin{matrix} \int_{(x,y)\in A}g(x,y,t)dx \\ \sum_{(x_i,y_i) \in A} TTC_i \end{matrix} \Bigg)
\end{equation}

With $A$ defined as in \eqref{eq:A}, $|R|$ is the region of interest in the image and $\gamma$ is a gain that modules the strength of the repulsive force.

\subsection{Road Potential Field} \label{sec:rpf}

Around the vehicle, the road boundaries should act like a barrier that prevents the car from departing from the lane. Then, the road potential field is designed in a way that it presents local maxima at the road boundaries and local minima in the center of the road, as that is the preferred position of the vehicle.

\begin{figure}
    \centering
    \includegraphics[width=\linewidth]{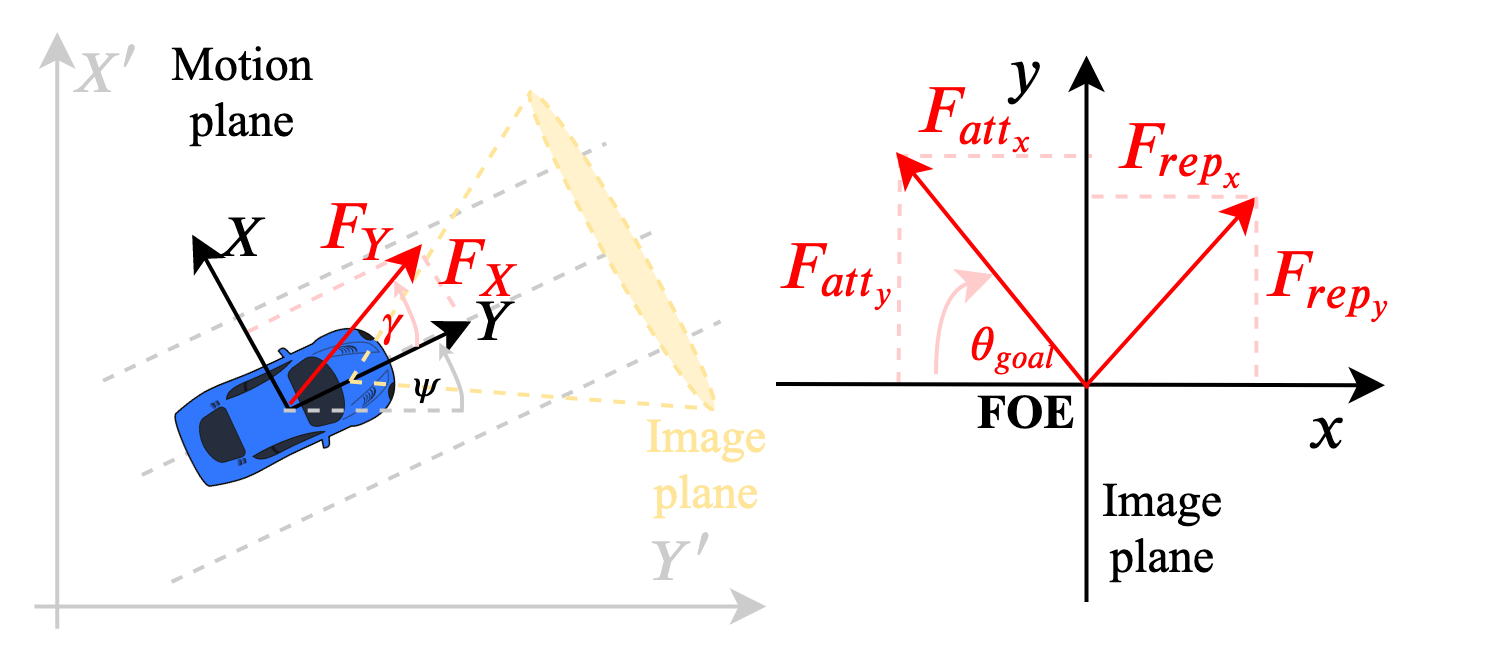}
    \caption{Motion plane coordinates are $(X,Y)$, image plane coordinates lie in the vertical plane $(x,y)$}
    \label{fig:coordveh}
    \end{figure}

The road edges are computed using the sparse optical flow. Then, a modified version of the Morse Potential Field is used for the design (\cite{snapper2018model,raksincharoensak2016vehicle}). Now, we will consider the local coordinates of the vehicle instead of the image plane, as seen in \ref{fig:coordveh}. The local coordinates form the plane $XY$, which differ from the image plane coordinates $xy$.

 Eq. \eqref{eq: str_tot} shows the expression for the total potential field ($U_s$) used for the straight segments of the road as the sum of the right ($U_{s_r}$) and left ($U_{s_l}$) lane potential fields.
\begin{equation}\label{eq: str_tot}
    U_s = U_{s_r} + U_{s_l}
\end{equation}

Where:
\begin{equation}\label{eq:str0_right}
        U_{s_r} = A \left( 1-e^{-b(y-y_r)} \right)^2,\quad U_{s_l} = A \left( 1-e^{b(y-y_l)} \right)^2
\end{equation}
        

In the above expressions, $A$ and $b$ are the depth and a parameter based on the variance of the repulsive potential field respectively. $y$ is the current position of the vehicle (considering the vehicle moves parallel to the $X$-axis), $y_r$ and $y_l$ are the relative distances to the right and left lane boundaries respectively. We consider that the vehicle moves in a four lane road, then $y_r = 5.25 (m)$ and $y_l = -8.75 (m)$ for a total road width of $14 (m)$ and a preferred position of the vehicle in the second lane from the right.

The computation of the right and left potential fields is designed as follows \cite{snapper2018model}:
\begin{equation}\label{eq:str1_right}
        U_{s_r} = A \left ( 1-e^{-b sign(y-y_r)\sqrt{\left( \frac{y-b_y}{m}-(x+\delta x) \right)^2+\left( y_r-y \right) ^2}}  \right)^2
\end{equation}
\begin{equation}\label{eq:str1_left}
        U_{s_l} = A \left ( 1-e^{b sign(y-y_l)\sqrt{\left( \frac{y-b_y}{m}-(x+\delta x) \right)^2+\left( y_l-y \right) ^2}}  \right)^2
\end{equation}

Where $x$ is the current $x$ position, $b_y$ is the y-intercept, $\delta x>0$ is a small number, $m$ is the slope of the line perpendicular to the lane center, and the rest of the variables are the same as before. In this case, both $y_r$ and $y_l$ are computed as follows:
\begin{flalign} 
        & y_r = c_2(x+\delta x)^n+c_1(x+\delta x)+c_{0_r}\\
        & y_l = c_2(x+\delta x)^n+c_1(x+\delta x)+c_{0_l}
\end{flalign}

In the previous equations,$c_{0_r}$ and $c_{0_l}$ represent the distance to the right lane marking from the center of the lane to the right and left respectively. The parameter $c_1$ is chosen as zero, and $c_2$ and $n$ will be chosen accordingly if the road is straight or curved.

To decide the curvature of the road, the position of the FOE is taken into account. If it is located in the center of the captured frame, then it is considered that the road is still straight for the next few seconds. In that case, $c_2 = 0.005$ and $n = 1$. If the FOE is located to the right or to the left, it is considered that the road is curved for the next few seconds. In that case, the choice of parameters $c_2 = \pm 5e-6$ (for left and right curve respectively) and $n = 2$ have given good results. 

Then, to compute the slope:
\begin{equation}\label{eq:m}
    m = -\frac{1}{2c_2(x+\delta x)+c_1}
\end{equation}

As seen in Eq. \eqref{eq:m}, $\delta x$ is needed to avoid $m$ to go to infinity when $x = 0$. Then, to compute the y-intercept of the line perpendicular to the lane center $b_y$:
\begin{equation}
    b_y = y_r-m(x+\delta x)
\end{equation}

Table \ref{table:apf} shows the chosen values for the parameters.

\begin{table}[]
\centering
\caption{Parameters used for APF implementation}
\label{table:apf}
\begin{tabular}{@{}lccc@{}}
\toprule
\multicolumn{1}{c}{\textbf{Description}} & \textbf{Symbol} & \textbf{Value} & \textbf{Unit} \\ \midrule
APF depth                                & A               & 0.5            &   -            \\
Parameter based on the variance          & b               & 1              &  -             \\
Parameter ($\pm$ curved road left/right)             & $c_2$           & 5e-6           &  -             \\
Parameter (straight road)                & $c_2$           & 0.005          & -              \\
Parameter                                & $c_1$           & 0              &  -             \\
Parameter (left,right side APF)               & $(c_{0_l},c_{0_r})$       & $(-8.75, 5.25)$          & (m)           \\
Small longitudinal offset                & $\delta x$      & 1.0e-10        & (m)           \\ \bottomrule
\end{tabular}
\end{table}

\begin{figure}[h!]
\centering
    \includegraphics[width=0.8\linewidth]{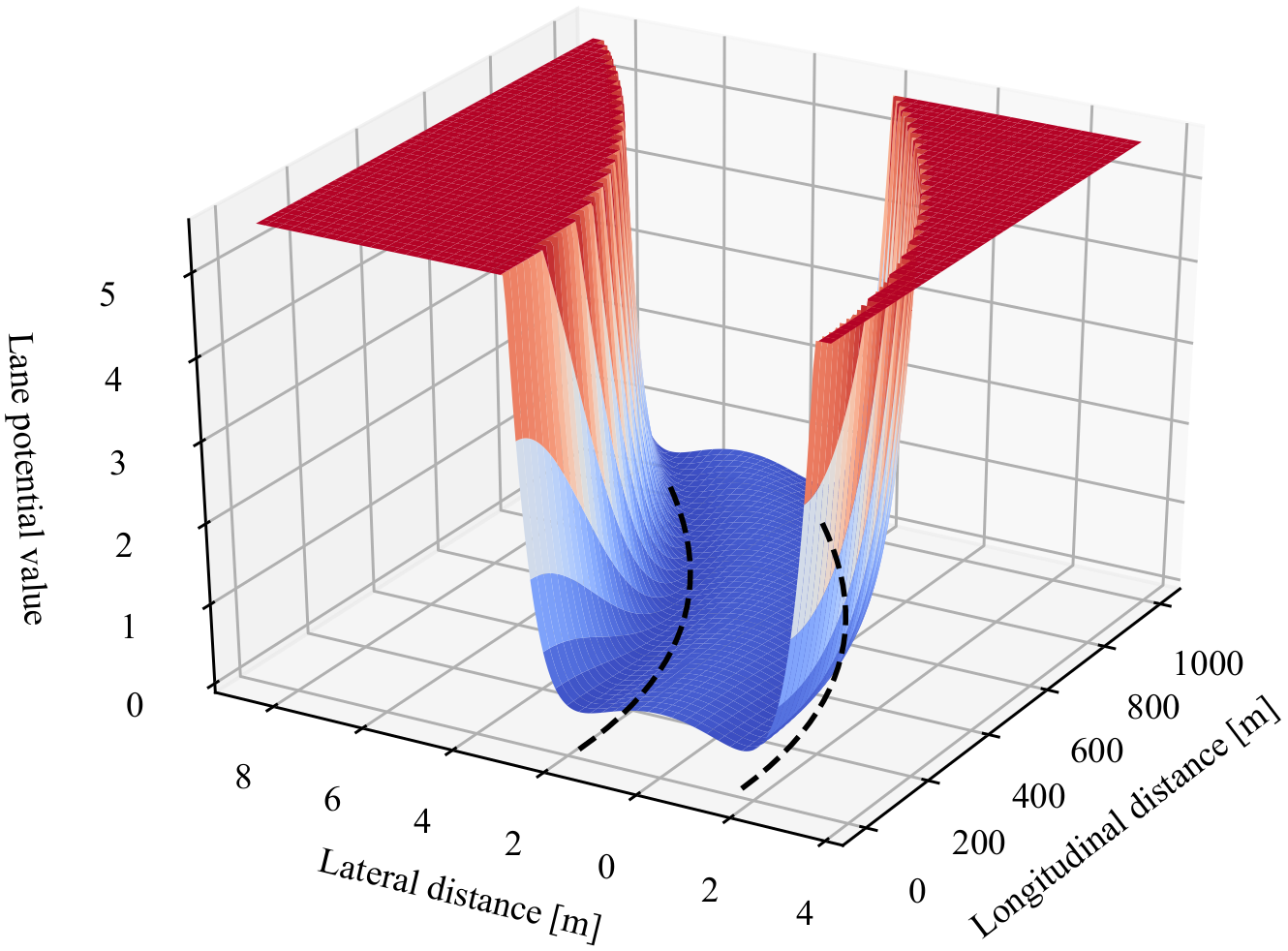}
    \caption{Angled view}
    \label{fig:curved}
  \end{figure}

The designed potential field is shown in Fig. \ref{fig:curved} for a one lane only left-curved road.

Finally, define the repulsive road force: $\textbf{F}_{rep_{rd}}=(F_{rep_{rd_X}},F_{rep_{rd_Y}})$:

\begin{equation}
    \textbf{F}_{rep_{rd}} = \nabla (U_{s_r} + U_{s_l})
\end{equation}

\section{Total Potential Field} \label{sec:tpf}

We consider the forces in the image plane and in the motion plane, seen in Fig. \ref{fig:coordveh} and developed in Sections \ref{sec:tpf}, \ref{sec:opf} and \ref{sec:rpf}.

\begin{flalign}\label{eq:ft}
        &F_{X_T}=F_{att_x}-F_{rep_x}-\lambda_X F_X\\
        &F_{Y_T}=F_{att_y}-F_{rep_y}-\lambda_Y F_Y
\end{flalign}

Where $\lambda_X,\lambda_Y>0$ are chosen appropriately to weigh the potential field from the motion plane with respect to the image plane. Ultimately, to find the force in the global coordinates:

\begin{equation}
    \Bigg( \begin{matrix} F_{X'}\\F_{Y'}  \end{matrix} \Bigg) = 
    \Bigg( \begin{matrix} \cos \psi && \sin \psi \\ -\sin \psi && \cos \psi \end{matrix} \Bigg)
    \Bigg( \begin{matrix}  F_{X_T}\\F_{Y_T} \end{matrix} \Bigg)
\end{equation}

Where $\psi$ is the orientation of the vehicle (or yaw angle) with respect to the global coordinates.

\section{Gradient Tracking Sliding Mode Controller (GTSMC)}

\subsection{Model of the vehicle}

The simplified bicycle kinematic model \cite{ozguner2011autonomous} of the vehicle is used for the design of the controller. Considering that only the front wheels can be steered, the model is:

\begin{flalign}\label{eq:kinematic}
        &\dot{x}=v\cos(\psi+\beta)\\
        &\dot{y}=v\sin(\psi+\beta)\\
        &\dot{\psi}=\frac{1}{l_f+l_r}v\cos\beta\tan\delta_f\\
        &\beta=arctan \left (\frac{l_r\tan\delta_f}{l_f+l_r} \right )\\
        &\dot{v}=a
\end{flalign}

\begin{figure}[h!]
    \centering
    \includegraphics[width=0.8\linewidth]{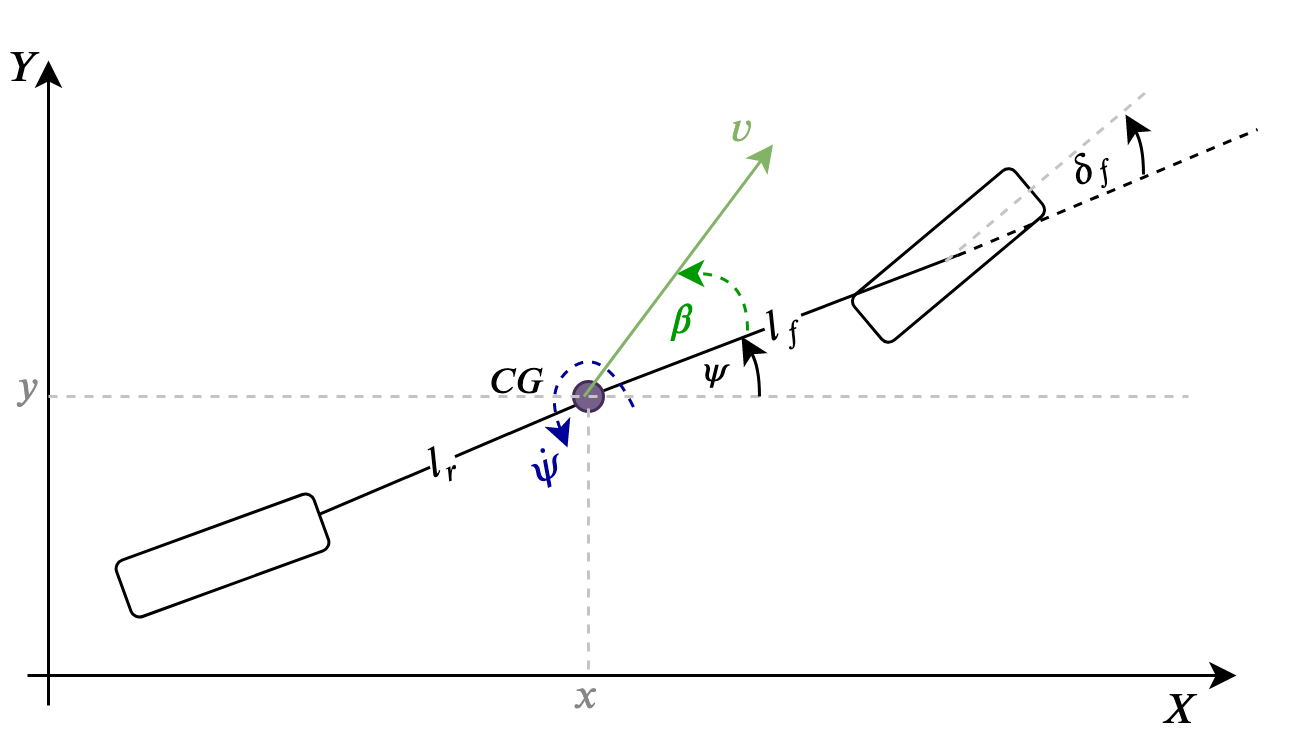}
    \caption{Kinematic bicycle model of a vehicle}
    \label{fig:kin_model}
\end{figure}

Where $x,y$ are the global $X,Y$-axis coordinates respectively, the yaw angle is represented by $\psi$ is the yaw angle (orientation of the vehicle with respect to the global X-axis and $\beta$ is the vehicle slip angle. The speed is $v:|$\textbf{v}$|$ and acceleration is $a$, in the same direction of the velocity \textbf{v}. The control inputs are the front steering angle $\delta_f$ and the acceleration $a$. This model is shown in Fig. \ref{fig:kin_model}.

\subsection{Controller design}

A gradient tracking sliding mode controller is chosen for the lateral controller \cite{guldner1995tracking,ferrara2007gradient}. The objective of the controller is to force the motion of the system $\dot{X}=f({X},u)$ to stay within a "sliding manifold" \cite{utkin2009sliding}. Since there is no preset trajectory, the gradient of the visual potential field is used to obtain an orientation reference \cite{ferrara2007gradient}.

Let $p=(x,y)$ be the position of the $CG$ of the vehicle. The motion direction against the gradient of the artificial potential field were already obtained in Section \ref{sec:tpf}. Then, for each point $p$ a continuous trajectory, called \textit{gradient line} is obtained. Thus, the desired orientation of the vehicle corresponds to:
\begin{equation}\label{phid}
    \psi_d(p)=atan\frac{F_{Y'}(p)}{F_{X'}(p)}
\end{equation}

Where $\psi_d(p) \in [-\pi,\pi[$. Then, we can find the rotational manifold $s_r$:
\begin{equation}
    s_r(p,t)=c_r\psi_e(p,t)+\dot{\psi}_e(p,t)
\end{equation}

Where $c_r>0$ is a constant, $\psi_e(p,t)=\psi(t)-\psi_d(p)$, with $\psi_d(p)$ from \eqref{phid}. 

Using the rotational manifold, we model the front wheel steering actuator as an integrator with constraints $|\delta|\leq \delta_0$,  $|u|\leq u_0$, and: \cite{guldner1995tracking}:
\begin{equation}
    \dot{\delta_f}=u
\end{equation}

\begin{equation}
    u=-u_0sign(s_r)
\end{equation}

When the amplitudes chosen are large enough it has been proven that the controller reaches the sliding manifold in finite time. Details are in \cite{utkin2009sliding}. 

A longitudinal manifold $s_t$ is used for the longitudinal controller with a constant $c_l>0$:
\begin{equation}
    s_l(p,t)=c_l v(p,t)-v_d(p,t)
\end{equation}

For the longitudinal control, the following simple sliding mode controller is chosen:
\begin{equation}
        a=-a_0sign(s_l)
\end{equation}

\section{Experiment Setup}

The experiment consists of two parts: first we use a driving simulator to extract images and obtain the proposed control and compare its performance against a PID. Then, we use a set of real images collected along with their control, and use our approach to predict the steering and throttle.

\subsection{Synthetic images}\label{sec:synthetic}

Synthetic images were extracted from Carla driving simulator \cite{dosovitskiy2017carla} because of its realistic 3D environments. 


The optical flow pyramidal implementation uses a size of the search window equal to $(25, 25)$ pixels and the accuracy threshold is $\epsilon=0.03$. The input image $I$ has the size $640 \times 480$ pixels, and three levels of the pyramid are used.

Table \ref{table:sim} shows the selected simulation parameters. The controlled variables are the throttle $a$ and steering angle $\delta$. These parameters accept normalized values: $a \in [0,1]$ for acceleration, $a \in [-1,0]$ for deceleration and $\delta \in [-1,1]$. The vehicle stops when it reaches certain predefined location in the map.

\begin{table}[]
\centering
\caption{Simulator setup}
\label{table:sim}
\begin{tabular}{@{}cc@{}}
\toprule
\textbf{Parameter}     & \textbf{Value}              \\ \midrule
Steering angle limits  & [-40\degree ,40\degree]     \\
Reference speed        & \textasciitilde 5.55m/s (20km/h)   \\
Weather                & ‘Clear Noon’, ‘Hard Rain’ \\
Scenario               & Highway Town04              \\
FPS                    & \textasciitilde60          \\
Window simulation size & 640 $\times$ 480 pixels           \\ \bottomrule
\end{tabular}
\end{table}

\begin{figure}[h!]
\includegraphics[width=\linewidth]{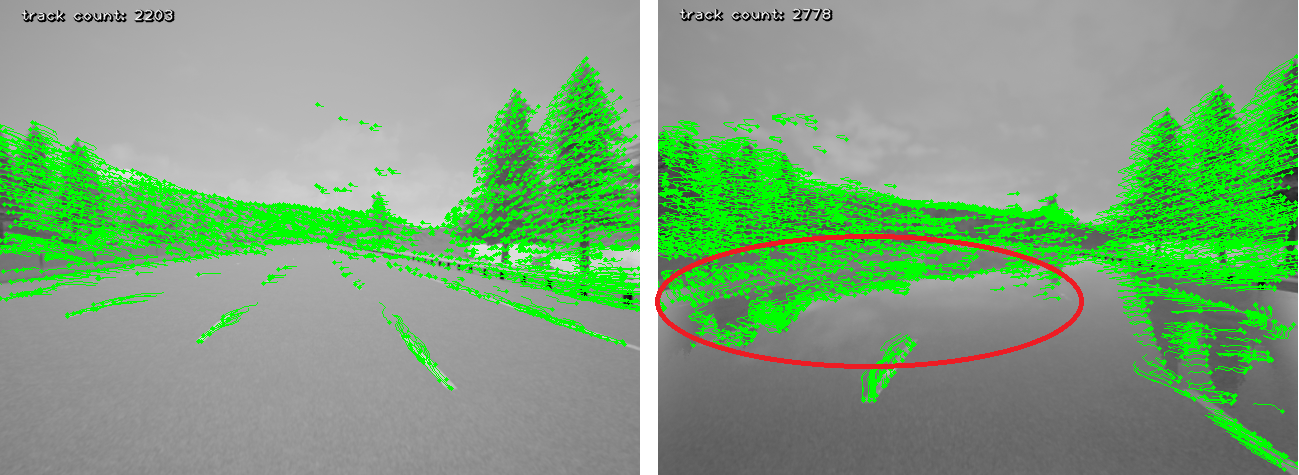}
\caption{Sparse optical flow in clear (left) and rainy (right) weather}
\label{fig:of}
\end{figure}

Two different weathers were tested in the same route, with and without obstacles on the road. This comparison is interesting because the rain makes the road look wet, almost like a mirror that reflects the objects in the scene, as seen in Fig. \ref{fig:of}. We evaluate how the vehicle behaves in both environments and present the path in terms of the global coordinates.

\begin{figure}[h!]
\centering
\includegraphics[width=0.8\linewidth]{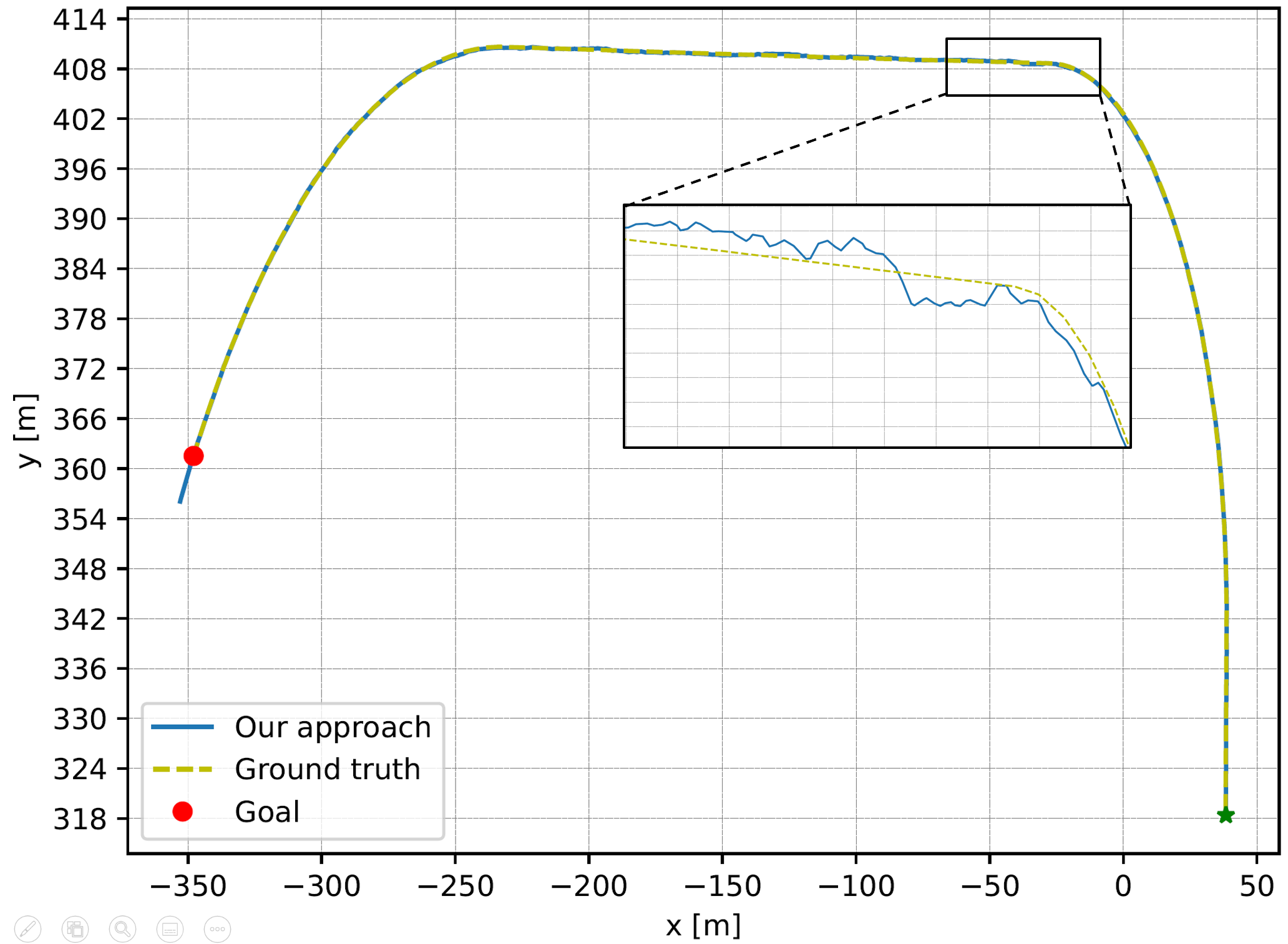}
\caption{Vehicle path with no obstacles in clear weather}
\label{fig:noon_noobs}
\end{figure}

The "ideal" route is defined via waypoints available in the simulator and it is expected that the vehicle be able to recreate a similar path. Fig. \ref{fig:noon_noobs} shows the path comparison when there are no obstacles and the weather is good. In the zoomed out plot it looks that the vehicle has followed perfectly the path, but once we zoom in, it is noticeable that there are deviations from the ideal path. Since the restrictions of the road considered that the whole 4 lanes of the road are allowable path, this behavior is expected.

\begin{figure}[h!]
\centering
\includegraphics[width=0.8\linewidth]{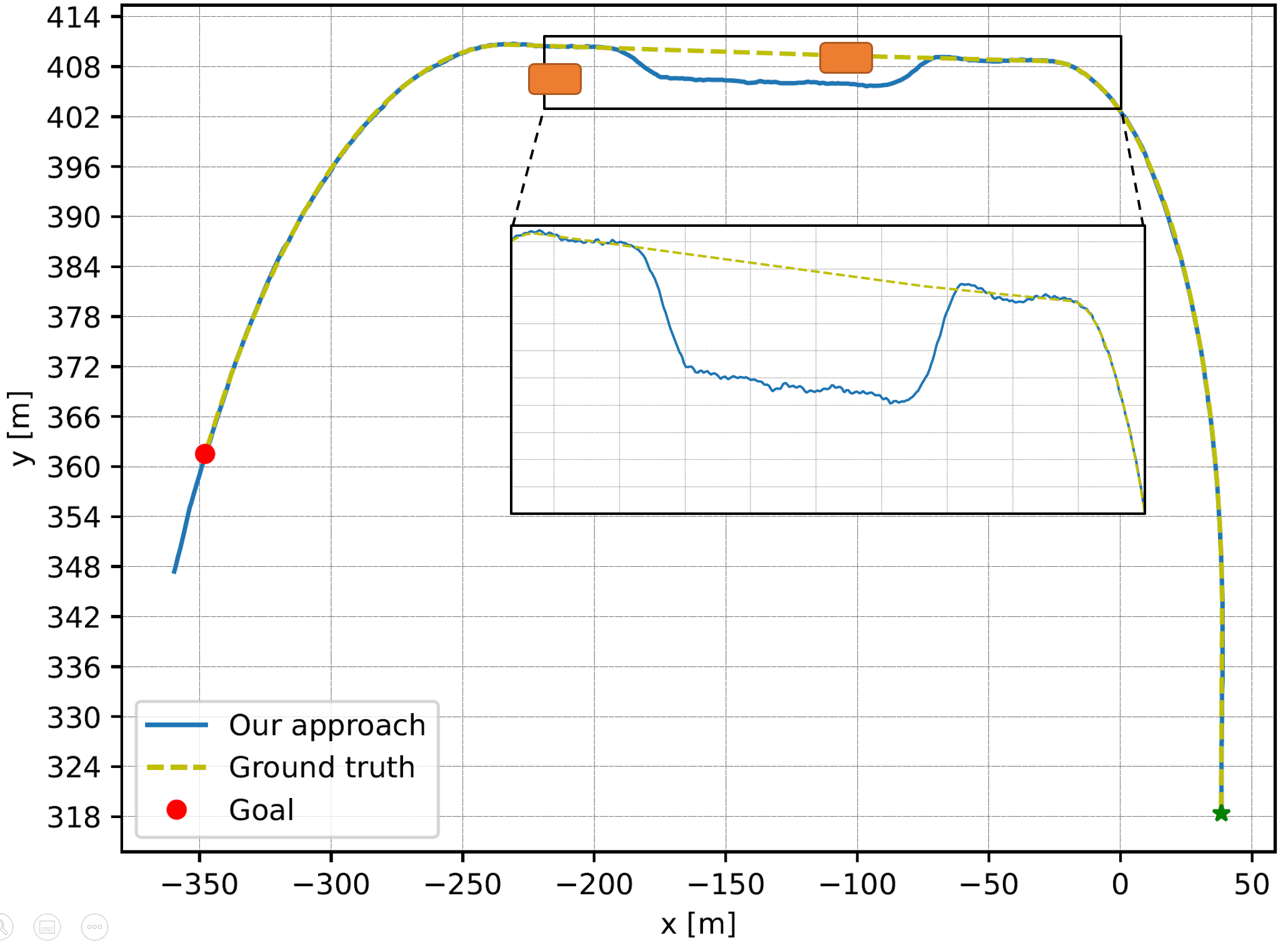}
\caption{Vehicle path with obstacles in clear weather, obstacles at $(-188.3,406.4)$ and $(-95.6,409.4.4)$}
\label{fig:noon_obs}
\end{figure}

By inserting obstacles in the previous weather, we get a disturbed path from the controlled vehicle, as seen in Fig. \ref{fig:noon_obs}. The visual potential field makes the vehicle change its direction to avoid both the obstacle and the lane boundaries.

\begin{figure}[h!]
\centering
\includegraphics[width=0.8\linewidth]{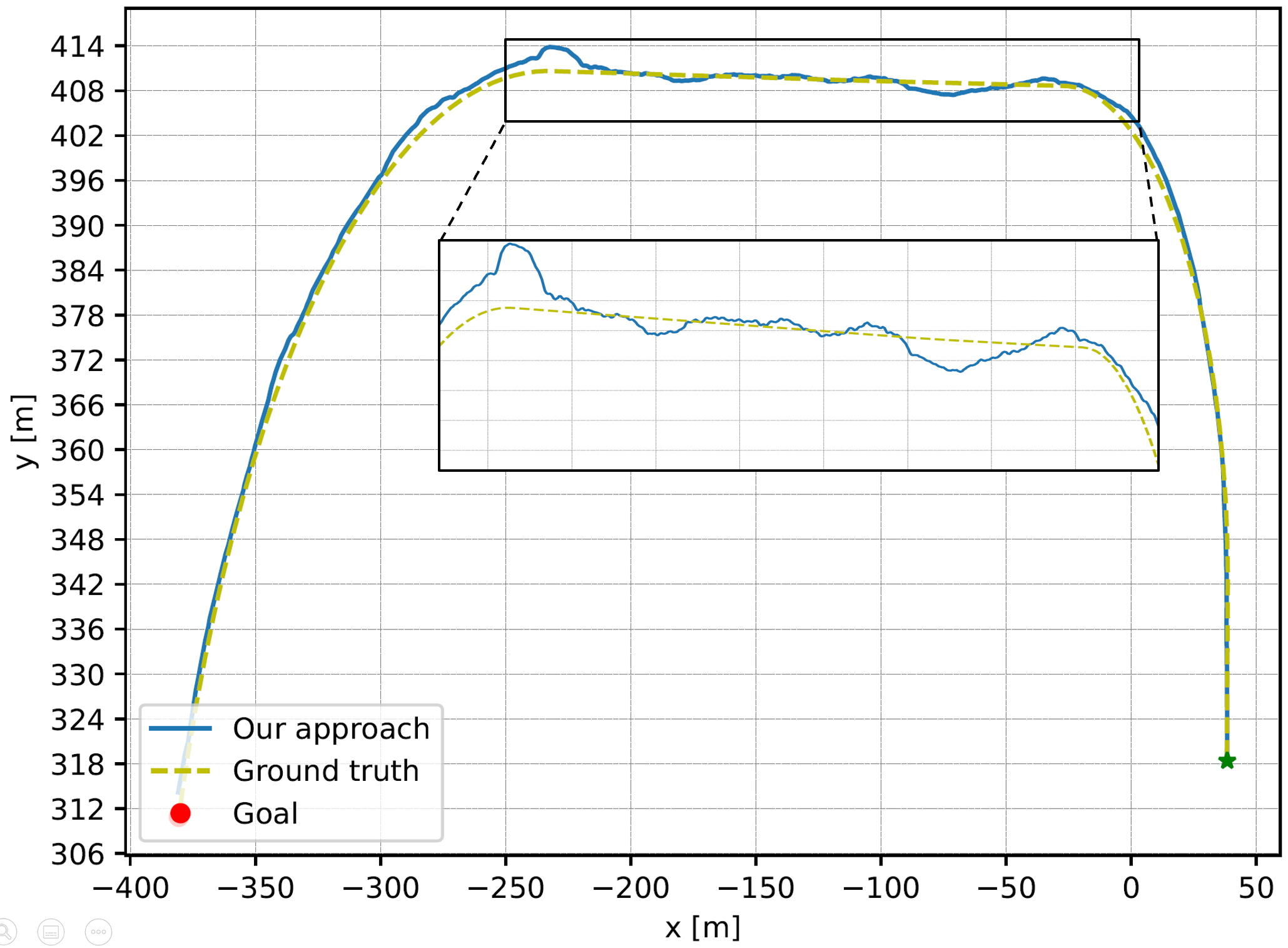}
\caption{Vehicle path with no obstacles in rainy weather}
\label{fig:rain_noobs}
\end{figure}

When there is rain in the scene, the reflection on the road causes disturbances in obtaining the obstacle map ($O_t(x,y)$), hence the path described by the vehicle is more jerky than when the weather is clear. Fig. \ref{fig:rain_noobs} shows this motion.

\begin{figure}[h!]
\centering
\includegraphics[width=0.8\linewidth]{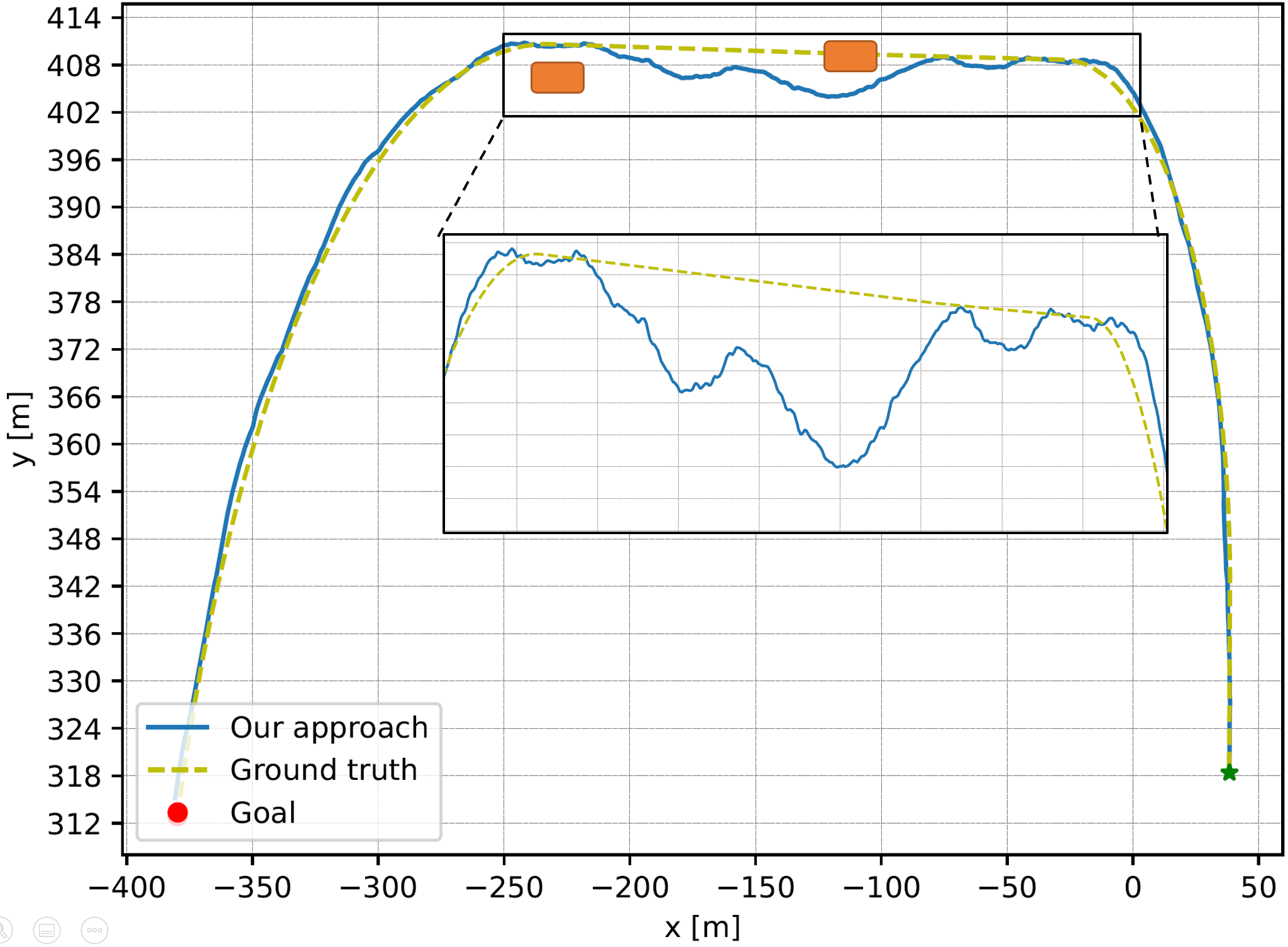}
\caption{Vehicle path with obstacles in rainy weather, obstacles at $(-188.3,406.4)$ and $(-95.6,409.4.4)$}
\label{fig:rain_obs}
\end{figure}

Inserting obstacles in the last situation makes it more challenging for the vehicle. We observe a very jerky motion in Fig. \ref{fig:rain_obs}. This behavior is expected, but nonetheless, the vehicle reaches its destination. 

It is worth noting that the vehicle was tested in a curved and a straight road, for which we designed precisely the road potential field. In case of having more complicated types of roads, it is necessary to generalize our method. So far, as long as we are taking a horizon of $t = 5$ seconds ($\approx 27.5 m$ at $20 km/h$) we are able to generate an estimated road potential field using the position of the FOE, as explained in Section \ref{sec:rpf}.


It is expected that the vehicle motion with obstacles would be jerkier. The disturbance in the obtained flow and potential field is reflected in the motion of the vehicle, seen in Fig. \ref{fig:rain_obs}. Here it is more noticeable that the vehicle cannot stay on its own lane, due in part to the fact that it is avoiding only the edges of the whole road, not only of its lane. This change was necessary to provide enough room to perform lane change (obstacle evasion).

Next, we compute a feasible path from the starting point to the end point using waypoints over the road, and use a longitudinal and lateral PID controller for the path tracking. We use the output of these controllers to compare the throttle and steering values obtained by our method.
We divide the route into obstacle and no obstacle section, and summarize the results from this experiment in Table \ref{tab:simres}. 

Whereas the prediction results for throttle and steering are not perfect, they are a good approximation that allow the average mean square error ($\overline{MSE}=\frac{1}{n} \sum_{i=1}^n (y_i-\hat{y}_i)^2+(x_i-\hat{x}_i)^2$) with respect to the original planned trajectory to be less than $1 (m)$ in average, which is one-third of the average US highway lane width ($\sim 3.7 (m)$).

\begin{table}[h!]
\centering
\caption{Simulator results comparison table, prediction accuracy ($\%$) for throttle $a$ and steering $\delta$ with respect to PID, mean square error for desired trajectory.}
\label{tab:simres}
\resizebox{\columnwidth}{!}{%
\begin{tabular}{@{}c|ccc|ccc@{}}
\toprule
\multicolumn{1}{l|}{\textbf{Weather}} & \multicolumn{3}{c|}{\textbf{No obstacles}} & \multicolumn{3}{c}{\textbf{With obstacles}} \\ \midrule
\textbf{} & \textbf{$a \%$} & \textbf{$\delta \%$} & \textbf{$ \overline{MSE}_{XY} (m)$} & \textbf{$a \%$} & \textbf{$\delta \%$} & \textbf{$\overline{MSE}_{XY} (m)$} \\
\textbf{Cloudy} & 82.34 & 72.14 & 0.42 & 78.51 & 71.67 & 0.74 \\
\textbf{Rain} & 78.45 & 67.12 & 0.97 & 70.76 & 65.45 & 1.34 \\ \bottomrule
\end{tabular}%
}
\end{table}


\subsection{Real Images Dataset}

Our approach is also tested on a set of real images that were collected alongside with their GPS, control and state information. The images show the front camera view of a vehicle in three weathers: cloudy, light rainy and rainy.

\begin{figure}[h!]
\includegraphics[width=\linewidth]{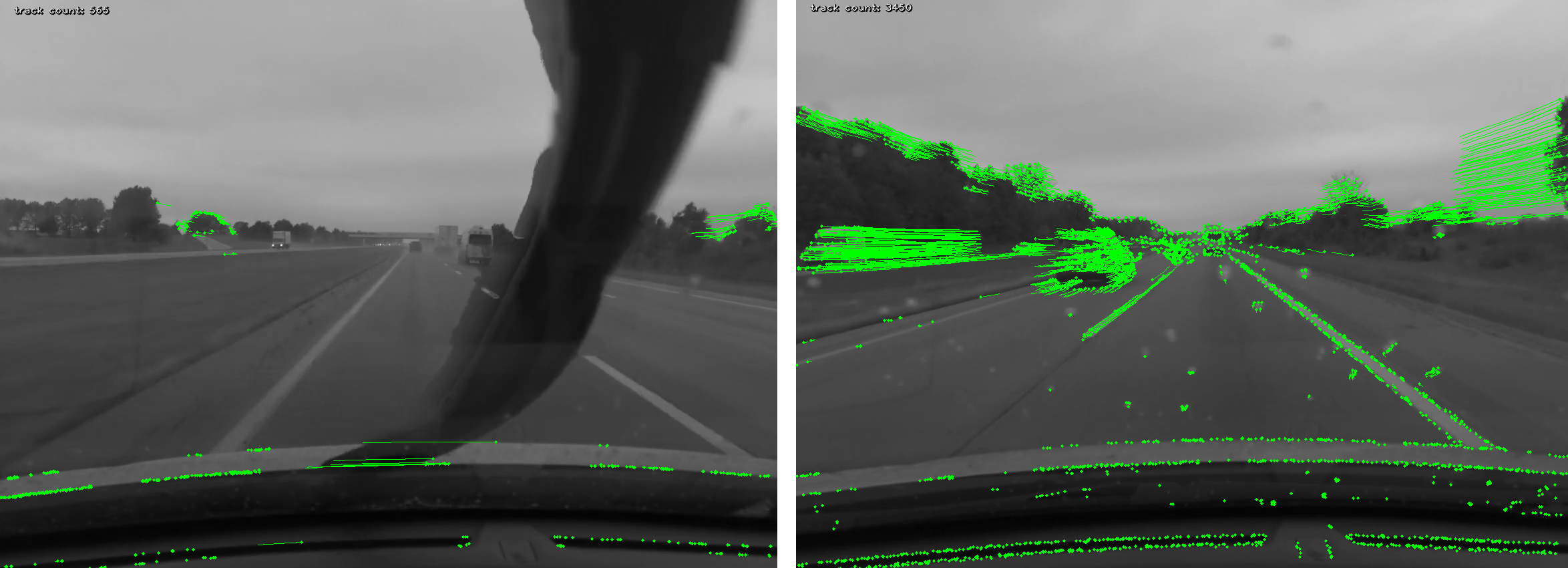}
\caption{Sparse optical flow when using wipers (left) and when droplets fall on the windshield (right)}
\label{fig:real}
\end{figure}

Analyzing the images, we notice that whenever it starts raining, our optical flow algorithm starts tracking the water droplets in the windshield. Furthermore, feature tracking is interrupted when the wipers are activated, see Fig. \ref{fig:real}. Identifying and removing these objects is important for this method to work, and is posed as a possible future contribution. However, we used the exact same pipeline as for the simulator synthetic images to evaluate its performance \textit{as is} .

We have analyzed each given route and divided them into obstacle free routes and routes with obstacles, to do a similar comparison as in Section \ref{sec:synthetic}. Then for each route, and by using the GPS coordinates information, we have selected an end point for our algorithm to represent the global minimum. The initial speed for the vehicle model is set according to the information provided by the dataset for each route segment.

We ran the images through the proposed pipeline and predicted the throttle $a$ and the steering angle $\delta$. We compare these results against the real throttle and steering angle from the dataset, and summarize in Table \ref{tab:realres}.

\begin{table}[]
\centering
\caption{Results comparison table for real dataset, prediction accuracy ($\%$) for throttle $a$ and steering $\delta$.}
\label{tab:realres}
\begin{tabular}{c|cc|cc}
\hline
\multicolumn{1}{l|}{\textbf{Weather}} & \multicolumn{2}{c|}{\textbf{No obstacles}} & \multicolumn{2}{c}{\textbf{With obstacles}} \\ \hline
\textbf{} & \textbf{$a \%$} & \textbf{$\delta \%$} & \textbf{$a \%$} & \textbf{$\delta \%$} \\
\textbf{Cloudy} & 69.43 & 64.92 & 67.12 & 62.34 \\
\multicolumn{1}{l|}{\textbf{Light rain}} & 58.27 & 55.87 & 55.31 & 54.23 \\
\textbf{Rain} & 56.12 & 48.51 & 50.97 & 46.12 \\ \hline
\end{tabular}%
\end{table}

As would be expected, the prediction accuracy for real images drops ($\sim50 - 60\%$), but still remains a promising result considering that the optical flow output was not modified. These results would increase with a tailored tuned optical flow with some removal methods to avoid the effect of droplets or wiper blades. 



\section{CONCLUSIONS}

This work showed that it is possible to obtain a visual potential field from the optical flow information from a monocular camera. The novelty of this work consists on the formulation of the potential field for both the obstacles and the road boundaries and applying it to control a vehicle. 



This visual based navigation method is less computationally expensive than learning based techniques, but at the same time, it allows to capture the features of dynamically changing environment. For this reason, it can serve as a baseline for comparison with both classical and learning based approaches.

Optical flow has its own limitations, such as being inaccurate for large motions, when there is occlusion and for strong illumination changes. Even when optical flow methods can be highly accurate in synthetic scenes, it is well known that its estimation on natural scenes can be a problem and require computationally expensive techniques to be solved, which contrasts with the benefits of low computation expense for the optical flow itself. This was shown in the set of real data presented in the experimental section.

Future work includes the comparison of this technique with learning based and classical ones, not only in performance, but in formulation complexity and execution time. Additionally, the method should be tested in more extensive sets of real data and ideally in a drive-by-wire vehicle to test not only the prediction, but actual performance. 





\section*{ACKNOWLEDGMENT}

This work was funded by the United States Department of Transportation
under award number 69A3551747111 for Mobility21: the National University
Transportation Center for Improving Mobility.\\
Any findings, conclusions, or recommendations expressed herein are those
of the authors and do not necessarily reflect the views of the United
States Department of Transportation, Carnegie Mellon University, or
The Ohio State University.

\bibliographystyle{IEEEtran}
\bibliography{root}

\begin{thebibliography}{10}
\providecommand{\url}[1]{#1}
\csname url@samestyle\endcsname
\providecommand{\newblock}{\relax}
\providecommand{\bibinfo}[2]{#2}
\providecommand{\BIBentrySTDinterwordspacing}{\spaceskip=0pt\relax}
\providecommand{\BIBentryALTinterwordstretchfactor}{4}
\providecommand{\BIBentryALTinterwordspacing}{\spaceskip=\fontdimen2\font plus
\BIBentryALTinterwordstretchfactor\fontdimen3\font minus
  \fontdimen4\font\relax}
\providecommand{\BIBforeignlanguage}[2]{{%
\expandafter\ifx\csname l@#1\endcsname\relax
\typeout{** WARNING: IEEEtran.bst: No hyphenation pattern has been}%
\typeout{** loaded for the language `#1'. Using the pattern for}%
\typeout{** the default language instead.}%
\else
\language=\csname l@#1\endcsname
\fi
#2}}
\providecommand{\BIBdecl}{\relax}
\BIBdecl

\bibitem{bonin2008visual}
F.~Bonin-Font, A.~Ortiz, and G.~Oliver, ``Visual navigation for mobile robots:
  A survey,'' \emph{Journal of intelligent and robotic systems}, vol.~53,
  no.~3, p. 263, 2008.

\bibitem{ohnishi2008visual}
N.~Ohnishi and A.~Imiya, ``Visual navigation of mobile robot using optical flow
  and visual potential field,'' in \emph{International Workshop on Robot
  Vision}.\hskip 1em plus 0.5em minus 0.4em\relax Springer, 2008, pp. 412--426.

\bibitem{miao2017optical}
H.~Miao and Y.~Wang, ``Optical flow based obstacle avoidance and path planning
  for quadrotor flight,'' in \emph{Chinese intelligent automation
  conference}.\hskip 1em plus 0.5em minus 0.4em\relax Springer, 2017, pp.
  631--638.

\bibitem{huang2006visual}
W.~H. Huang, B.~R. Fajen, J.~R. Fink, and W.~H. Warren, ``Visual navigation and
  obstacle avoidance using a steering potential function,'' \emph{Robotics and
  Autonomous Systems}, vol.~54, no.~4, pp. 288--299, 2006.

\bibitem{camus1995calculating}
T.~Camus and T.~Camus, \emph{Calculating time-to-contact using real-time
  quantized optical flow}.\hskip 1em plus 0.5em minus 0.4em\relax US Department
  of Commerce, National Institute of Standards and Technology, 1995.

\bibitem{gern2002vision}
A.~Gern, R.~Moebus, and U.~Franke, ``Vision-based lane recognition under
  adverse weather conditions using optical flow,'' in \emph{Intelligent Vehicle
  Symposium, 2002. IEEE}, vol.~2.\hskip 1em plus 0.5em minus 0.4em\relax IEEE,
  2002, pp. 652--657.

\bibitem{lieb2005adaptive}
D.~Lieb, A.~Lookingbill, and S.~Thrun, ``Adaptive road following using
  self-supervised learning and reverse optical flow.'' in \emph{Robotics:
  science and systems}, 2005, pp. 273--280.

\bibitem{yoshimoto1995automatic}
K.-i. Yoshimoto, M.~Sakatoh, M.~Takeuchi, and H.~Ogawa, ``An automatic steering
  control algorithm using optical flow,'' \emph{JSAE review}, vol.~16, no.~2,
  pp. 165--169, 1995.

\bibitem{okafuji2015development}
Y.~Okafuji, T.~Fukao, and H.~Inou, ``Development of automatic steering system
  by modeling human behavior based on optical flow,'' \emph{Journal of Robotics
  and Mechatronics}, vol.~27, no.~2, pp. 136--145, 2015.

\bibitem{souhila2007optical}
K.~Souhila and A.~Karim, ``Optical flow based robot obstacle avoidance,''
  \emph{International Journal of Advanced Robotic Systems}, vol.~4, no.~1,
  p.~2, 2007.

\bibitem{chang2016experimental}
R.~Chang, R.~Ding, M.~Lin, D.~Meng, Z.~Wu, and M.~Hang, ``An experimental
  evaluation of balance strategy based obstacle avoidance,'' in \emph{Control,
  Automation, Robotics and Vision (ICARCV), 2016 14th International Conference
  on}.\hskip 1em plus 0.5em minus 0.4em\relax IEEE, 2016, pp. 1--6.

\bibitem{guldner1995tracking}
J.~Guldner, V.~Utkin, H.~Hashimoto, and F.~Harashima, ``Tracking gradients of
  artificial potential fields with non-holonomic mobile robots,'' in
  \emph{American Control Conference, Proceedings of the 1995}, vol.~4.\hskip
  1em plus 0.5em minus 0.4em\relax IEEE, 1995, pp. 2803--2804.

\bibitem{ferrara2007gradient}
A.~Ferrara and M.~Rubagotti, ``Gradient tracking based second order sliding
  mode control of a wheeled vehicle,'' in \emph{Control Conference (ECC), 2007
  European}.\hskip 1em plus 0.5em minus 0.4em\relax IEEE, 2007, pp. 3810--3817.

\bibitem{snapper2018model}
E.~Snapper, ``Model-based path planning and control for autonomous vehicles
  using artificial potential fields,'' Master's thesis, TU Delft, 2018.

\bibitem{hamid2018}
S.~Y. Z. H. R.~P. Hamid, UZA, ``Collision avoidance performance analysis of a
  varied loads autonomous vehicle using integrated nonlinear controller,''
  \emph{PERINTIS eJournal}, vol.~8, no.~1, pp. 17---43, 2018.

\bibitem{raksincharoensak2016vehicle}
P.~Raksincharoensak, T.~Hasegawa, A.~Yamasaki, H.~Mouri, and M.~Nagai,
  ``Vehicle motion planning and control for autonomous driving intelligence
  system based on risk potential optimization framework,'' in \emph{The
  Dynamics of Vehicles on Roads and Tracks: Proceedings of the 24th Symposium
  of the International Association for Vehicle System Dynamics (IAVSD 2015),
  Graz, Austria, 17-21 August 2015}.\hskip 1em plus 0.5em minus 0.4em\relax CRC
  Press, 2016, p. 189.

\bibitem{capito2019optical}
L.~J. Capito~Ruiz, ``Optical flow-based artificial potential field generation
  for gradient tracking sliding mode control for autonomous vehicle
  navigation,'' Master's thesis, The Ohio State University, 2019.

\bibitem{shi1993good}
J.~Shi and C.~Tomasi, ``Good features to track,'' Cornell University, Tech.
  Rep., 1993.

\bibitem{bouguet2001pyramidal}
J.-Y. Bouguet, ``Pyramidal implementation of the affine lucas kanade feature
  tracker description of the algorithm,'' \emph{Intel Corporation}, vol.~5, no.
  1-10, p.~4, 2001.

\bibitem{otsu1979threshold}
N.~Otsu, ``A threshold selection method from gray-level histograms,''
  \emph{IEEE transactions on systems, man, and cybernetics}, vol.~9, no.~1, pp.
  62--66, 1979.

\bibitem{ozguner2011autonomous}
U.~Ozguner, T.~Acarman, and K.~Redmill, \emph{Autonomous ground
  vehicles}.\hskip 1em plus 0.5em minus 0.4em\relax Artech House, 2011.

\bibitem{utkin2009sliding}
V.~Utkin, J.~Guldner, and J.~Shi, \emph{Sliding mode control in
  electro-mechanical systems}.\hskip 1em plus 0.5em minus 0.4em\relax CRC
  press, 2009.

\bibitem{dosovitskiy2017carla}
A.~Dosovitskiy, G.~Ros, F.~Codevilla, A.~Lopez, and V.~Koltun, ``Carla: An open
  urban driving simulator,'' \emph{arXiv preprint arXiv:1711.03938}, 2017.

\end{thebibliography}

\end{document}